\documentclass[letterpaper, 10pt, conference]{IEEEtran}
\IEEEoverridecommandlockouts
\usepackage{cite}
\usepackage{amsmath,amssymb,amsfonts}
\usepackage{algorithmic}
\usepackage{graphicx}
\usepackage{textcomp}
\usepackage{xcolor}
\def\BibTeX{{\rm B\kern-.05em{\sc i\kern-.025em b}\kern-.08em
    T\kern-.1667em\lower.7ex\hbox{E}\kern-.125emX}}

\usepackage{lineno,hyperref}
\modulolinenumbers[5]
\usepackage{framed,multirow}
\usepackage{float}
\usepackage{booktabs}
\usepackage{placeins}
\usepackage[ruled,vlined]{algorithm2e}
\usepackage{setspace}
\usepackage{cuted}
\usepackage{balance}
\usepackage{enumitem}

\newcommand{\etal}{\textit{et al.~}}
\usepackage{soul}
\usepackage{todonotes}

\begin{document}

\title{Multiscale Analysis for Improving Texture Classification\\
}

\author{Steve T. M. Ataky$^{1}$, Diego Saqui$^{2}$, Jonathan de Matos$^{1}$, Alceu S. Britto Jr.$^{3}$ and Alessandro L. Koerich$^{1}$
\thanks{}
\thanks{$^{1}$Steve T. M. Ataky, Jonathan de Matos, and Alessandro L. Koerich are with \'{E}cole de Technologie Sup\'{e}rieure, University of Qu\'{e}bec, H3C 1K9, Montr\'{e}al, QC, Canada.
        {\url{steve.ataky@nca.ufma.br}, \url{jonathan.de-matos.1@ens.etsmtl.ca}, \url{alessandro.koerich@etsmtl.ca}}}
\thanks{$^{2}$Diego Saqui is with Instituto Federal do Sul de Minas Gerais, Muzambinho, MG, Brazil.
{\url{diego.saqui@ifsuldeminas.edu.br}}}
\thanks{$^{3}$ Alceu S. Britto Jr. is with Pontificia Universidade Catolica do Paran\'a, Curitiba, PR, 80215-901, Brazil.
{\url{alceu.britto@pucpr.br}}}%
}


\maketitle
\begin{abstract}
Information from an image occurs over multiple and distinct spatial scales. Image pyramid multiresolution representations are a useful data structure for image analysis and manipulation over a spectrum of spatial scales. This paper employs the Gaussian-Laplacian pyramid to treat different spatial frequency bands of a texture separately. First, we generate three images corresponding to three levels of the Gaussian-Laplacian pyramid for an input image to capture intrinsic details. Then we aggregate features extracted from gray and color texture images using bio-inspired texture descriptors, information-theoretic measures, gray-level co-occurrence matrix features, and Haralick statistical features into a single feature vector. Such an aggregation aims at producing features that characterize textures to their maximum extent, unlike employing each descriptor separately, which may lose some relevant textural information and reduce the classification performance. The experimental results on texture and histopathologic image datasets have shown the advantages of the proposed method compared to state-of-the-art approaches. Such findings emphasize the importance of multiscale image analysis and corroborate that the descriptors mentioned above are complementary. 
\end{abstract}

\begin{IEEEkeywords}
Texture analysis, Texture characterization, Image pyramid, Multiscale image analysis.
\end{IEEEkeywords}


\section{Introduction}
\label{sec:intro}
The definition of texture has different flavors in the literature on computer vision. A common one considers texture as changes in the image intensity that form specific repetitive patterns~\cite{tuceryan1993texture}. These patterns may result from the physical properties of the object's surface (roughness) that provide different types of light reflection. A smooth surface reflects the light at a defined angle (specular reflection), while a rough surface reflects it in all directions (diffuse reflection). Although texture recognition is easy for human perception, it is not the case with automatic procedures where this task sometimes requires complex computational techniques. In computer vision, texture analysis is of a notable role and whose basis is the extraction of relevant information from an image to characterize its texture. This process involves a set of algorithms and techniques. Since humans' perception of texture is not affected by rotation, translation, and scale changes, any numerical image texture characterization should be invariant to those aspects and any monotonic transformation in pixel intensity. 

Several image texture analysis methods have been developed in the last decades, exploring different properties to characterize texture information within an image~\cite{simon2018review,liu2019bow, reviewtexture}. Some of the classical and novel approaches developed are gray-level co-occurrence matrix (GLCM)~\cite{haralick1973textural}, local binary patterns (LBP)~\cite{pietikainen2011computer}, Haralick descriptors~\cite{haralick1979statistical}, Markov random fields~\cite{cross1983markov}, wavelet transform~\cite{arivazhagan2003texture}, Gabor texture discriminator among others, and more recently, the bio-inspired texture descriptor (BiT)~\cite{ataky2021novel}. 

Besides, convolutional neural networks (CNNs) have held the attention of researchers due to their effectiveness in tasks such as object detection. However, they are not too suitable for texture classification~\cite{texturecnn01,MatosBOK19}. For instance, Andrearczyk and Whelan~\cite{texturecnn01} proposed a simple texture CNN (T-CNN) architecture that pools an energy measure at its last convolution layer and discards the shape information usually learned by classic CNNs. Unfortunately, the trade-off between accuracy and complexity is not favorable even though the results are encouraging. Likewise, other texture CNN architectures with moderate performance on texture classification are presented in \cite{MatosBOK19,fujieda2017wavelet,Vriesman2019}. 

Texture descriptors are also quite popular in medical image analysis, particularly in histopathologic images (HIs), due to the different textures found in HIs. Areas representing high/low concentration of nuclei and stroma, for instance. For this reason, several researchers have been investigating a broad spectrum of textural descriptors for HI classification, such as GLCM, LBP, Gabor, and Haralick~\cite{electronics10050562}. 

Even though the descriptors mentioned above have shown an important discriminative power individually, combining them may be a promising strategy to provide a representation based on several intrinsic textural properties. With this in mind, we evaluate the combination of texture descriptors such as the BiT~\cite{ataky2021novel}, information-theoretic measures~\cite{shannon1948mathematical}, GLCM~\cite{haralick1973textural}, and Haralick features~\cite{haralick1979statistical} to improve the performance of texture classification. To this end, the feature vector representing an image is formed by concatenating features provided by different descriptors individually. The rationale is to compensate for the possible loss when using a single technique to give the textural description. This process is carried out over a spectrum of spatial scales representation using Gaussian-Laplacian (GLP), a useful data structure for image analysis and manipulation. 

The contribution of this paper is threefold: (i) the combination of BiT descriptor, information theory, Haralick, and GLCM features for texture characterization; (ii) a better discriminating ability while using color and gray-scale features on different categories of images; (iii) a method for texture classification that represents the state-of-the-art on challenging datasets.

This paper is organized into five sections. The multiresolution concepts used in this work are presented in Section~\ref{sec:related}. Section~\ref{sec:method} presents the proposed approach. The experimental design, the datasets used, our results and related discussion are presented in Section~\ref{sec:experiment}. The last section presents the conclusion and future work prospects. 

\section{Multiresolution Analysis}
\label{sec:related}
Humans usually see regions of similar textures, colors, or gray levels that combine to form objects when looking at an image. If objects are small or have low contrast, it may be necessary to examine them in high resolution; a coarser view is satisfactory if they are large or have high contrast. If both types of objects appear in an image, it can be helpful to analyze them in multiple resolutions. Changing the resolution can also lead to the creation, deletion, or merging of image features. Also, there is evidence that the human visual system processes visual information in a multiresolution way~\cite{blakemore1969existence}. Besides, sensors can provide data in various resolutions, and multiresolution algorithms for image processing offer advantages from a computational point of view and are generally robust. 

\begin{figure}[htpb!]
\begin{minipage}[b]{0.32\linewidth}
  \centering
  \centerline{\includegraphics[width=2.4cm]{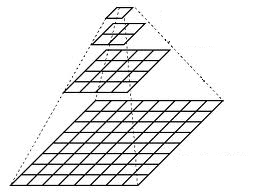}}
  \centerline{\scriptsize (a) Pyramid representation}\medskip
\end{minipage}
\begin{minipage}[b]{.33\linewidth}
  \centering
  \centerline{\includegraphics[width=2.4cm]{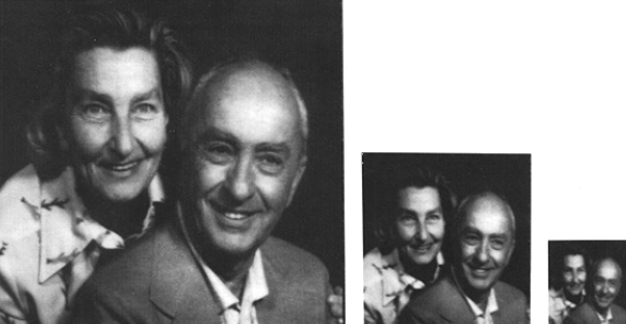}}
  \centerline{\scriptsize (b) Gaussian pyramid}\medskip
\end{minipage}
\begin{minipage}[b]{0.33\linewidth}
  \centering
  \centerline{\includegraphics[width=2.4cm]{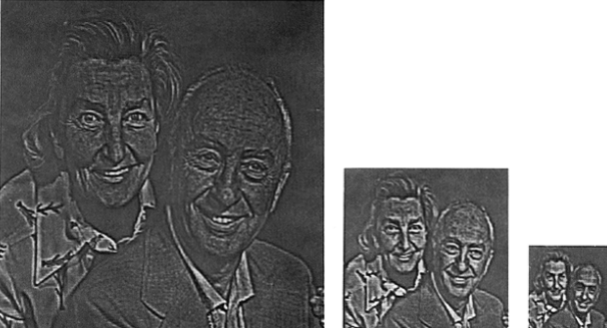}}
  \centerline{\scriptsize (c) Laplacian Pyramid}\medskip
\end{minipage}
\caption{An example of Gaussian and Laplacian Pyramids from the same input image. (b) First three levels of Gaussian pyramid; (c) First three levels of Laplacian pyramid. Images from \cite{Ataky2020}.}
\label{fig:pyramids}
\end{figure}

When analyzing an image, it can sometimes be helpful to break it down into separate parts so that there is no loss of information. The pyramid theory provides ways to decompose images at multiple levels of resolution. It considers a collection of representations of an image in different spatial resolutions, stacked on top of each other, with the highest resolution image at the bottom of the stack and subsequent images appearing over it in descending order of resolution. This generates a pyramid-like structure, as shown in Fig.~\ref{fig:pyramids}(a). 

The traditional procedure for obtaining a lower resolution image is to perform low-pass filtering followed by sampling~\cite{jolion2012pyramid}. In signal processing and computer vision, pyramid representation is the main type of multiscale representation for computing image features on different scales. The pyramid is obtained by repeated smoothing and subsampling of an image. For generating the pyramid representation, different smoothing kernels have been brought forward, and the binomial one strikingly shows up as useful and theoretically well-founded~\cite{lindeberg2013scale}. 

Accordingly, for a bi-dimensional image, the normalized binomial filter may be applied (1/4, 1/2, 1/4) in most cases twice or even more along all spatial dimensions. The subsampling of the image by a factor of two follows, which leads to efficient and compact multilevel representation. There are low-pass and band-pass pyramid types~\cite{jolion2012pyramid}. 

To develop filter-based representations by decomposing images into information on multiple scales and extracting features/structures of interest from an image, the Gaussian pyramid (GP), the Laplacian pyramid (LP), and the wavelet pyramid are examples of the most frequently used pyramids. The GP (Fig.~\ref{fig:pyramids}) consists of low-pass filtered, reduced density, where subsequent images of the preceding level of the pyramid are weighted down using Gaussian average or Gaussian blur and scaled-down. The original image is defined as the base level. Formally speaking, assuming that $I(x,y)$ is a two-dimensional image, the GP is recursively defined as presented in \eqref{eq:GP}. 

\scriptsize
\begin{equation}
	\begin{aligned}
		G_0(x,y) =\begin{cases}
		    I(x,y), & \hspace{-1mm} \text{for } l = 0\\
		    \\
			\sum\limits_{m = -2}^{2} \sum\limits_{n = -2}^{2} w(m, n)G_{l-1}(2x+m, 2y+n), & \hspace{-1mm} \text{otherwise.}
		\end{cases}
	\end{aligned}
	\label{eq:GP}
\end{equation}
\normalsize

\noindent where $w(m, n)$ is a weighting function (identical at all levels) termed the generating kernel, which adheres to the following properties: separable, symmetric, and each node at level $n$ contributes the same total weight to nodes at level $l+1$. The pyramid name arose because the weighting function nearly approximates a Gaussian function. This pyramid holds local averages on different scales,  leveraged for target localization and texture analysis. 

Moreover, assuming the GP $[{I}_{0}, {I}_{1}, \dots, {I}_{k}]$, the LP is obtained by computing ${b}_{k} = {I}_{k} - E {I}_{k+1}$, where $ E {I}_{k+1}$ represents an up-sampled, smoothed version of ${I}_{k+1}$ of the same dimension. In the literature, LP is used for analysis, compression, and enhancement of images and graphics applications~\cite{jolion2012pyramid}. 

\vspace{-1mm}

The GP is used in this work for the multiresolution representation of the images right before the feature extraction.

\section{Proposed Approach}
\label{sec:method}
\begin{figure*}[htpb!]
	\centering
	\includegraphics[width=0.95\textwidth]{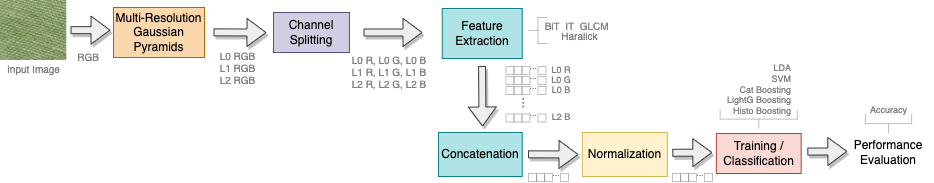}
	\caption[Methodology]{An overview of the proposed scheme.}
	\label{Fig:model}
\end{figure*}

This section describes how the proposed method integrates multiresolution analysis and multiple texture descriptors for texture classification. To this end, we put forward an architecture structured in five main stages as follows. Fig.~\ref{Fig:model} shows an overview of the proposed scheme, and Algorithm~\ref{algo:method} shows the steps of the first three stages. 

\vspace{-1mm}

\noindent{\it Multiresolution representation:} in this stage, for an input image, we generate three  {others} corresponding to {the} levels of the Gaussian pyramid (L0: original image, L1: first level of the pyramid, L2: second level of the pyramid). The purpose is to represent an input image in different resolutions to capture intrinsic details. 

\noindent{\it Channel Splitting:} each channel (R, G, B) is considered a separate input in this phase. The key reason behind the splitting channels is to exploit color information. Thus, we represent and characterize an input image in a given resolution by a set of local descriptors generated from the interaction of a pixel with its neighborhood inside a given channel (R, G, or B). 

\noindent{\it Feature Extraction:} after the channel splitting step, the images undergo feature extraction, which looks for intrinsic properties and discriminative characteristics. For each GP level of an image, we extract: BiT~\cite{ataky2021novel} (14-dimensional), Shannon entropy and multi-information, aka total information (3-dimensional), Haralick~\cite{haralick1973textural} (13-dimensional), and GLCM~\cite{haralick1973textural} (6-dimensional) from each channel. Images are then represented by the concatenation of different measurements organized as feature vectors (Algorithm~\ref{algo:method}). For simplicity, we name the resulting descriptor \textit{Three-in-One} (TiO). It is worth mentioning that images were converted in gray-scale for GLCM and Haralick measurement to be extracted but remained color for extracting the bio-inspired indices. After feature extraction and before concatenation, the feature vectors have 126, 117, 54, and 18 dimensions for BiT, Haralick, GLCM, and Shannon entropy and multi-information, respectively. Thereby, after concatenation, we have a 315-dimensional feature vector. 

\noindent{\it Normalization:} because test points simulate real-world data, we split the data into training and test sets and perform min-max normalization on the training data. Subsequently, we used the same min-max values to normalize the test set. 
 
\noindent{\it Training/Classification:} we have a normalized feature vector comprising the concatenation of features extracted from each resolution and color channel from the previous stage. This texture representation is used to train one monolithic and three ensemble-based models. The linear discriminant analysis (LDA) is used for the monolithic classification model, while ensemble models are created using the histogram-based algorithm for building gradient boosting ensembles of decision trees (HistoB), the light gradient boosting decision trees (LightB), and the CatBoost\footnote{https://catboost.ai/}, which is an efficient implementation of the gradient boosting algorithm. 

\begin{algorithm}[htpb!]
\scriptsize
\linespread{1.0} 
\SetAlgoLined
\KwResult{feature descriptor}
 \textbf{1. Read} a RGB image file \;
 \textbf{2. Generate 3 multiresolution levels with Gaussian pyramids \textbf{L0, L1, L2}}\;
 \bf 3. For each level (L0, L1, L2) of Gaussian pyramids \;
 \textbf{\quad 3.1. Separate} the image in channels \textbf{R}, \textbf{G}, \textbf{B}\;
 \textbf{\quad 3.2. Convert} \textbf{R}, \textbf{G}, and \textbf{B} to grayscale images (for GLCM and Haralick only)\;
 \textbf{\quad 3.3. Compute} Bio-inspired indices, Shannon entropy and multi-information, Haralick, and GLCM of \textbf{R}, \textbf{G}, \textbf{B} \;
 \textbf{\quad 3.4. Concatenate} these values into a single vector (315-dimensional)\;
 \textbf{4. Repeat} steps 1 to 4 for all images of the dataset
 \caption{FeatureExtractionProcedure}
 \label{algo:method}
\end{algorithm}


\section{Experimental Results and Discussion}
\label{sec:experiment}
The experimental protocol considers five datasets, three of which are texture image collections, and two are composed of medical images. 
The KTH-TIPS dataset\footnote{\url{https://www.csc.kth.se/cvap/databases/kthtips/download.html}} contains a collection of 810 color texture images of 200$\times$200 pixel of resolution. The images were captured at nine scales, under three different illumination directions and three different poses, with 81 images per class. Seventy percent of images are used for training, while the remaining 30\% are used for testing. The Outex TC 00011-r~\cite{ojala2002outex} has a total of 960 (24$\times$20$\times$2) images of illuminant Inca. The training set consists of 480 (24$\times$20) images and the test set of 480 (24$\times$20) images. The Amsterdam Library of Textures (ALOT) is a color image collection of 250 (classes) rough textures. The authors systematically varied {the} viewing angle, illumination angle, and illumination color for each material so that they could capture the sensory variation in object recordings. CRC is a dataset of colorectal cancer histopathology images of 5,000$\times$5,000 pixels that were patched into 150$\times$150 pixels images and labeled according to eight types of structures. The total number of images is 625 per structure type, resulting in 5000 images. Finally, the BreakHis dataset comprises 9,109 microscopic images of breast tumor tissue collected from 82 patients using different magnification factors (40$\times$, 100$\times$, 200$\times$, and 400$\times$). {Currently}, it contains 2,480 benign and 5,429 malignant samples (700$\times$460 pixels, 3-channel RGB, 8-bit depth in each channel, PNG format). 
    

We split each dataset into training (70\%) and test (30\%) sets to evaluate all the classifiers on the same experimental condition. In the experiments, the full feature vector generated by concatenating all descriptors (TiO) is used for training a monolithic classifier using the LDA algorithm and ensembles using the HistoB, LightB, and CatBoost algorithm. 

\subsection{Experiments with Texture Datasets}
Table~\ref{tab:acc_all} presents the results of the proposed method (TiO), BiT, GLCM, and Haralick over Outex, KTH-TIPS, and ALOT datasets. For each dataset, we present the results obtained by combining all the descriptors (TiO) and the results obtained with each descriptor individually. The main purpose is not only to verify the effectiveness of TiO but also the complementary descriptors in the study. The proposed method achieved the best average accuracy of 100\% on the Outex dataset with LDA and CatBoost classifiers. On the ALOT dataset, the proposed method achieved an average accuracy of 98.48\% with the LDA classifier. Again, TiO outperformed all individual descriptors. Finally, on the KTH-TIPS dataset, the proposed method achieved the best average accuracy of 100\% with the LDA classifier. Moreover, TiO outperformed all individual descriptors. 

\begin{table}[htpb!]
\setlength{\tabcolsep}{3pt}
\caption{Average accuracy (\%) on the test set of Outex, ALOT and KTH-TIPS datasets. The best accuracy for each dataset is shown in boldface. The best result for each texture descriptor is marked with $^*$.}
\label{tab:acc_all}
\footnotesize
\centering
 \begin{tabular}{|r|r|c|c|c|c|} 
 \hline
 & \bfseries  Texture  & \multicolumn{4}{c|}{\bfseries Classification Algorithms}\\ 
\cline{3-6}
\bfseries  Dataset & \bfseries Descriptors & \bfseries HistoB$^1$ & \bfseries LightB$^2$ & \bfseries LDA & \bfseries CatBoost$^3$\\  
 \hline
\multirow{3}{*}{}
  \multirow{4}{*}{Outex} & TiO      & {\bfseries 100.0} & {\bfseries 100.0} & {\bfseries 100.0} & {\bfseries 100.0}\\
   & BiT      & 99.92  & 99.69 & 100.0$^*$ & 99.84\\
   & GLCM      & 99.69$^*$  & 97.99 & 95.91 & 99.30  \\
   & Haralick      & 98.84 & 99.53 & 99.38 & 99.84$^*$  \\
 \hline
 \multirow{3}{*}{}
  \multirow{4}{*}{ALOT} & TiO          &  41.50 & 43.07 & {\bfseries 98.48} & 93.47\\
   & BiT         & 10.71 & 11.52 & 68.14 & 74.90$^*$  \\
   & GLCM        & 11.28 & 14.52 & 66.61$^*$ & 64.23 \\
   & Haralick      & 18.52 & 20.47 & 86.57$^*$ & 83.23 \\
 \hline
 \multirow{3}{*}{}
   \multirow{4}{*}{KTH-TIPS} & TiO &  98.53 & 97.20 & {\bfseries 100.0} & 98.91 \\
   & BiT     & 96.29 & 96.70 & 99.58$^*$ & 97.53  \\
   & GLCM    & 89.71  & 90.12 & 87.65 & 92.01$^*$   \\
   &Haralick    & 95.88 & 96.29 & 98.35$^*$ & 96.11  \\
\hline
\multicolumn{6}{l}{$^1$max\_bins=10, max\_iter=1500. $^{2,3}$n\_estimators=1500.}
\end{tabular}
\end{table}
\vspace{-1mm}

Different works have used the Outex dataset for texture classification. For instance,~\cite{kth2} introduced an approach based on a rotation-invariant LBP, achieving an accuracy of 96.26\% with a $k$-NN classifier. \cite{kth3} presented a rotation-invariant, impulse noise resistant, and illumination-invariant approach based on a local spiking pattern. This approach achieved an accuracy of 86.12\% with a neural network but is not extended for color textures, and it requires several input parameters. \cite{ataky2021novel} introduced a bio-inspired texture descriptor based on biodiversity (species richness and evenness) and taxonomic measures. In the latter, the image is represented as an abstract model of an ecosystem where species diversity and richness and taxonomic distinctiveness measures are extracted. Such a texture descriptor is invariant to rotation, translation, and scale and achieved the accuracy of 99.88\% with an SVM. Table~\ref{tab:acc_outex_relatedWorks} compares our best results with a few works that also used the Outex dataset. 

\begin{table}[htpb!]
\caption{Accuracy (\%) of the proposed method and related works on Outex dataset. The best result is marked with $^*$.}
\label{tab:acc_outex_relatedWorks}
\footnotesize
\centering
 \begin{tabular}{|c|l|c|} 
  \hline
\bfseries   Reference &\bfseries  Approach & \bfseries Accuracy  \\ 
  \hline
\multirow{3}{*}{\cite{ataky2021novel}} & GLCM & 95.52  \\
                             & Haralick & 96.92  \\
                             & BiT & 99.88  \\
\hline
\cite{kth2} & Shallow & 96.23  \\
\hline
\cite{kth3} & Shallow & 86.12  \\
\hline
TiO+LDA or TiO+CatBoost & Shallow & 100.0$^*$   \\
\hline
\end{tabular}
\end{table}
\vspace{-1mm}

Table \ref{tab:acc_alot_relatedWorks} presents a few works that also used the ALOT dataset. There was an improvement in the accuracy of nearly 1\% and 0.1\% compared with shallow and deep methods, respectively. Notwithstanding the difference is not significantly high, it is worth mentioning that the success of CNNs places reliance on the ability to leverage massive labeled datasets to learn high-quality representations. Nonetheless, data availability for a few domains may be restricted, and therefore CNNs become restrained from several fields. Moreover, some works evaluated small-scale CNN architectures, such as T-CNN and T-CNN Inception, with 11,900 and 1.5M parameters, respectively. Despite the reduced number of parameters and lower computational cost for training, both still required a large amount of training data to perform satisfactorily. Even some small architectures of comparable performance, such as MobileNets, EfficientNets, and sparse architectures resulting from pruning, still have many training parameters. For instance, the number of parameters of a MobileNet ranges between 3.5M and 4.2M, while EfficientNets range between 5.3M and 66.6M. GoogleNet, ResNet, and VGG generally need extensive training, and the number of hyperparameters and the computational cost is high. 

\begin{table}[htpb!]
\caption{Accuracy (\%) of shallow and deep approaches on the ALOT dataset. The best result is marked with $^*$.}
\label{tab:acc_alot_relatedWorks}
\footnotesize
\centering
 \begin{tabular}{|c|l|c|} 
  \hline
\bfseries   Reference & \bfseries Approach & \bfseries Accuracy \\ 
  \hline
  \cite{Alot2} & Shallow & 97.56  \\
                  \hline
  \cite{Alot3} & Shallow & 63.04   \\
                  \hline
\multirow{2}{*}{\cite{Alot4}} & ResNet50 & 75.68   \\
                & ResNet101 &  75.60  \\
                \hline
\multirow{7}{*}{\cite{Alot1}} & ResNet101 & 98.13  \\
                 & ResNet50 & 98.35  \\
                 & VGG VeryDeep19 & 94.93  \\
                 & VGG M128 & 85.56 \\
                 & GoogleNet & 92.65  \\
                 & VGG M1024 & 92.58  \\
                 & VGG M2048 & 93.30  \\
                                 \hline
 \cite{Alot5} & Shallow & 97.20   \\
                 \hline
 TiO+LDA & Shallow & 98.48$^*$   \\
 \hline
\end{tabular}
\end{table}
\vspace{-1mm}

Likewise, the KTH-TIPS dataset has been used to evaluate texture characterization and classification approaches. Hazgui \etal\cite{kth1} introduced an approach that integrates the genetic programming and the fusion of HOG and LBP features, which achieved an accuracy of 91.20\% with a $k$-NN classifier. Such an approach does not use color information and global features. \cite{kth4} put forth rotational and noise invariant statistical binary patterns, which reached an accuracy of 97.73\%, which is lower than the accuracy achieved by the proposed method of about 2.3\%. This approach is resolution sensitive and presents a high computational complexity. Qi \etal\cite{qi2013exploring} proposed a rotation-invariant multiscale cross-channel LBP (CCLBP) that encodes the cross-channel texture correlation. The CCLBP computes the LBP descriptors in each channel and three scales and computes co-occurrence statistics before concatenating the extracted features. Such an approach achieved an accuracy of 99.01\% for three scales with an SVM. Nevertheless, this method is not invariant to scale. Table~\ref{tab:acc_kth_relatedWorks} shows that the proposed approach outperforms other works that also used the KTH-TIPS. 

\begin{table}[htpb!]
\caption{Accuracy (\%) of the proposed method and related works on KTH-TIPS dataset. The best result is marked with $^*$.}
\label{tab:acc_kth_relatedWorks}
\footnotesize
\centering
 \begin{tabular}{|c|l|c|} 
  \hline
\bfseries   Reference & \bfseries Approach & \bfseries Accuracy\\ 
  \hline
    \multirow{3}{*}{\cite{ataky2021novel}} & GLCM & 86.83 \\
                             & Haralick & 94.89 \\
                             & BiT & 97.87 \\
    \hline
    \cite{kth1} & Shallow & 91.20 \\
    \hline
    \cite{kth4} & Shallow & 97.73 \\
    \hline
    \cite{qi2013exploring} & Shallow & 99.01 \\
    \hline
 TiO+LDA & Shallow & 100.0$^*$ \\
 \hline
\end{tabular}
\end{table}
\vspace{-1mm}

\subsection{Experiments with HI Datasets}
Table~\ref{tab:acc_crc} presents the accuracy achieved by monolithic classifiers and ensemble methods, both trained with the proposed method on the CRC dataset with TiO and BiT, GLCM, and Haralick, individually. The proposed approach provided its best accuracy of 94.71\%, with HistoB and LightB with all the features. The accuracy difference between TiO and the first best-related work is nearly 2.00\%, which corroborates the discriminating ability of our method. Additionally, compared to each descriptor that TiO is made up of, TiO outperformed each of them when employed individually. 

Table~\ref{tab:acc_crc_relatedWorks} compares the results of the proposed approach with some state-of-the-art works to assess its effectiveness. The results achieved by TiO on the CRC dataset have shown that the proposed approach works well on images with other structures apart from textures and with no need for data augmentation. Besides, such CNNs need to be trained with a large amount of data {which may} be prohibitive in fields such as medical imaging. 
 
\begin{table}[htpb!]
\caption{Accuracy (\%) of monolithic classifiers and ensemble methods with TiO and each descriptor employed individually on the CRC dataset.}
\label{tab:acc_crc}
\footnotesize
\centering
 \begin{tabular}{|r| c| c| c| c|} 
 \hline
         \bfseries Texture  & \multicolumn{4}{c|}{\bfseries Classification Algorithms}\\ 
\cline{2-5}
   \bfseries Descriptors & \bfseries HistoB & \bfseries LightB & \bfseries LDA & \bfseries CatBoost \\  
 \hline
TiO &\textbf{ 94.71} & \textbf{94.71} & 94.37 & 93.80  \\
 \hline
    BiT       & 93.22 & 92.61 & 86.67 & 92.92  \\
 \hline
    GLCM      & 85.98 & 86.97 & 81.79 & 87.20  \\
 \hline
    Haralick  & 91.62 & 91.31 & 92.38 & 91.39 \\
\hline
\end{tabular}
\end{table}
\vspace{-1mm}

\begin{table}[htpb!]
\caption{Average accuracy (\%) of shallow and deep approaches on the CRC dataset. The best results are marked with $^*$.}
\label{tab:acc_crc_relatedWorks}
\footnotesize
\centering
 \begin{tabular}{|c|c|l|c|} 
  \hline
  \bfseries Reference & \bfseries Approach & \bfseries 10-fold & \bfseries 5-fold \\ 
  \hline
\cite{CRC_kather2016multi} & Shallow & 87.40 & --\\
    \hline
 \cite{CRC_sarkar2017sdl}& Shallow & 73.60 &  -- \\
    \hline
 \cite{ataky2021novel} & Shallow &  92.96 & -- \\
    \hline
 {TiO+HistoB} & Shallow & 94.71$^*$ & 92.77$^*$ \\
    \hline
 \cite{CRC_wang2017histopathological} & CNN & -- & 92.60 \\
    \hline
 \cite{CRC_pham2017scaling} & CNN & -- & 84.00 \\
    \hline
 \cite{CRC_rkaczkowski2019ara} & CNN & 92.40 & 92.20 \\
 \hline
\end{tabular}
\end{table}
\vspace{-1mm}

Table~\ref{tab:acc_break_bal} shows the accuracy achieved by monolithic classifiers and ensemble methods trained with all the features of the proposed approach. For the BreakHis dataset, the HistoB classifier achieved the best accuracy for 40$\times$, 100$\times$ and 400$\times$ magnifications, and LightB for 200$\times$ magnification. 

Table~\ref{tab:acc_breakhis_relatedWorks} shows and compares the results achieved by the proposed approach using all the features with the state-of-the-art for the BreakHis dataset. One can note that the proposed approach achieved a considerable accuracy of 98.64\% with all the features for 40$\times$ magnification, which slightly outperforms the accuracy of both shallow and deep methods. The difference in accuracy between the proposed approach and the second-best method (CNN) is nearly 1\% for 40$\times$ magnification. Likewise, the proposed method achieved a considerable accuracy of 97.85\%, 98.76\% and 98.22\% for 100$\times$, 200$\times$, and 400$\times$, respectively, which slightly outperformed the second-best method with difference of 0.9\% for 100$\times$, 1.05\% for 200$\times$, and 1.0\% for 400$\times$ magnifications, respectively. 

We have also carried out the experiments with BiT, GLCM, and Haralick separately. For all the magnifications, TiO outperformed each descriptor with the maximum accuracy difference of 1.25\%, 0.48\%, 1.01\%, 1.5\% for 40$\times$, 100$\times$, 200$\times$, and 400$\times$ magnifications, respectively. Thus, combining the descriptors mentioned above increased accuracy by nearly 1\%. 

\begin{table}[htpb!]
\caption{Accuracy (\%) of monolithic classifiers and ensemble methods with TiO descriptor on balanced 8-classes image-level BreakHis dataset. The best result for each magnification is marked with $^*$.}
\label{tab:acc_break_bal}
\footnotesize
\centering
 \begin{tabular}{|c|c|c|c|c|} 
 \hline
         & \multicolumn{4}{c|}{\bfseries Classification Algorithms}\\ 
\cline{2-5}
\bfseries  Magnification    & \bfseries HistoB & \bfseries LightB & \bfseries LDA &\bfseries  CatBoost \\  
 \hline
40$\times$ & {98.64}$^*$ & 98.12 & 97.83 & 98.04   \\ 
 \hline
100$\times$ & {97.85}$^*$ & 97.50 & 96.81 & 96.67  \\
 \hline
200$\times$ & 97.62 & {98.76}$^*$ & 94.63 & 97.28  \\
 \hline
400$\times$ &{ 98.22}$^*$ & 98.06 & 94.92 & 96.98  \\
 \hline
\end{tabular}
\end{table}
\vspace{-1mm}

\begin{table}[htpb!]
\setlength{\tabcolsep}{5pt}
\caption{Accuracy (\%) of shallow and deep approaches on the BreakHis dataset. All works used the same data partitions for training and test. The best results are indicated with $^*$.}
\label{tab:acc_breakhis_relatedWorks}
\footnotesize
\centering
 \begin{tabular}{|c|c|c|c|c|c|} 
 \hline
 & & \multicolumn{4}{c|}{\bfseries Magnification}\\ 
\cline{3-6}
 \bfseries  Reference & \bfseries Method & \bfseries 40$\times$ & \bfseries 100$\times$ & \bfseries 200$\times$ & \bfseries 400$\times$\\
 \hline
 \cite{Spanhol2016} (LBP) & Shallow &  75.60 & 73.00  & 72.90 & 71.20\\
  \hline
\cite{Spanhol2016} (GLCM) & Shallow &74.70 & 76.80 & 83.40 & 81.70\\
 \hline
\cite{BreakHis_1}& Shallow &  88.30 & 88.30  & 87.10 & 83.40\\ 
 \hline
 \cite{ataky2021novel}& Shallow   &  { 97.50} & 96.80  & 95.80 & 95.20\\
  \hline
\cite{BrealHis_2}& CNN &  97.00 & {97.50} & {97.20} & {97.20} \\
 \hline
 \cite{BreakHis_3}& CNN &  92.80 & 93.90  & 93.70 & 92.90 \\
 \hline
 \cite{BreakHis_4}& CNN &  83.00 & 83.10  & 84.60 & 82.10 \\
 \hline
 \cite{BreakHis_5}& CNN &  90.00 & 88.40  & 84.60 & 86.10\\
  \hline
TiO+HistoB & Shallow   &  {98.64}$^*$ & {97.85}$^*$  & {98.76}$^*$ & { 98.22}$^*$\\
  \hline
\end{tabular}
\end{table}
\vspace{-1mm}

\section{Conclusions}
\label{sec:conclusion}
This research provided an important study regarding image analysis and manipulation over a spectrum of spatial scales and the complementarity of feature descriptors for texture classification. We stated that employing each descriptor individually may overlook relevant textural information, reducing the classification performance. Moreover, we exploited the pyramids' multiresolution representation as a useful data structure for analyzing and capturing intrinsic details from texture over a spectrum of spatial scales. To produce a feature vector from an image, we combined several descriptors that have proven to be discriminating for the classification, namely, the BiT, information-theoretic measures, GLCM, and Haralick descriptors, to extract gray-level and color features in different resolutions. Such a combination aimed to bring features that characterize the texture to the maximum extent and with some advantages such as rotation, permutation, and scale-invariant, reduced noise sensitivity, generic and high generalization ability, as it has provided effective performance for the real-world datasets. 

The proposed approach outperformed a few state-of-the-art shallow and deep methods. Moreover, the descriptors employed herein have proven to be complementary as their combination resulted in a better performance than using each one individually. However, some features may be redundant after the concatenation into a single feature vector, given the different resolutions, channels, and descriptors. That may cause a downfall in the classification performance. Furthermore, such a concatenation may also lead to the Hughes phenomenon, which explains why we have not included other descriptors such as LBP, HOG, etc. However, we will consider including other descriptors in our future studies. Finally, to circumvent the possible feature redundancy and cope with the increase in dimensionality, we will investigate the impact of incorporating a decision marker multi-objective feature selection. 





\bibliographystyle{IEEEtran}
\bibliography{refs}

\end{document}